\documentclass{article}
\usepackage{codefuse_tech_report}
\usepackage[colorlinks = true,
            linkcolor = blue,
            urlcolor  = blue,
            citecolor = blue,
            anchorcolor = blue]{hyperref}   
\usepackage{microtype}
\usepackage{xurl}
\usepackage{booktabs}
\usepackage{enumitem}
\usepackage{multicol}
\usepackage{CJKutf8}
\usepackage{siunitx}
\usepackage{floatflt}
\usepackage{graphicx}
\usepackage{wrapfig}
\usepackage{authblk}
\usepackage{lipsum}
\usepackage{algorithm}
\usepackage{algorithmicx}
\usepackage{algpseudocode}
\usepackage{subfigure}
\usepackage{multirow}
\usepackage{pifont}  

\algnewcommand{\LeftComment}[1]{\Statex \(\triangleright\) #1}

\usepackage{array}
\usepackage{amsmath}
\usepackage{amssymb}
\usepackage{mathtools}
\usepackage{amsthm}

\usepackage[capitalize,noabbrev]{cleveref}
\usepackage{adjustbox} 

\theoremstyle{plain}

\theoremstyle{definition}

\theoremstyle{remark}

\colmfinalcopy

\usepackage[utf8]{inputenc}
\usepackage[T1]{fontenc}
\usepackage{caption} 
\usepackage{adjustbox} 
\usepackage{fontawesome5}

\usepackage{color}
\usepackage[table]{xcolor}
\usepackage{soul} 

\definecolor{tred}{RGB}{251, 130, 132}
\definecolor{torange}{RGB}{247, 162, 116}
\definecolor{tyellow}{RGB}{251, 218, 140}
\definecolor{tgreen}{RGB}{127, 204, 181}
\definecolor{tblue}{RGB}{89, 177, 215}
\definecolor{insightblue}{RGB}{162, 210, 255}
\definecolor{questionred}{RGB}{255, 175, 204}

\title{F2LLM Technical Report: Matching SOTA Embedding Performance with 6 Million Open-Source Data}

\author{%
Ziyin Zhang$^{1,2}$
~~Zihan Liao$^{1}$
\\

\vspace{-6pt}
\bf
~~Hang Yu\thanks{Correspondence to: Hang Yu \textless hyu.hugo@antgroup.com\textgreater, Peng Di \textless dipeng.dp@antgroup.com\textgreater, Rui Wang \textless wangrui12@sjtu.edu.cn\textgreater.}~~$^{,1}$ 
~~Peng Di$^{*, 1}$
~~Rui Wang$^{*, 2}$

\vspace{10pt}
$^1$Ant Group\ \ \ $^2$Shanghai Jiao Tong University\\
\vspace{10pt}
\hspace{-10pt}\faGithub ~\href{https://github.com/codefuse-ai/CodeFuse-Embeddings/tree/main/F2LLM}{https://github.com/codefuse-ai/CodeFuse-Embeddings}\\
\hspace{-10pt}~~~~~~~~\includegraphics[width=1em,height=1em]{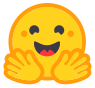} ~\href{https://huggingface.co/collections/codefuse-ai/codefuse-embeddings-68d4b32da791bbba993f8d14}{https://huggingface.co/collections/codefuse-ai/codefuse-embeddings}\\
}

\colmfinalcopy 

\begin{document}

\maketitle

\begin{figure}[h!]
    \centering
    \includegraphics[width=0.59\linewidth]{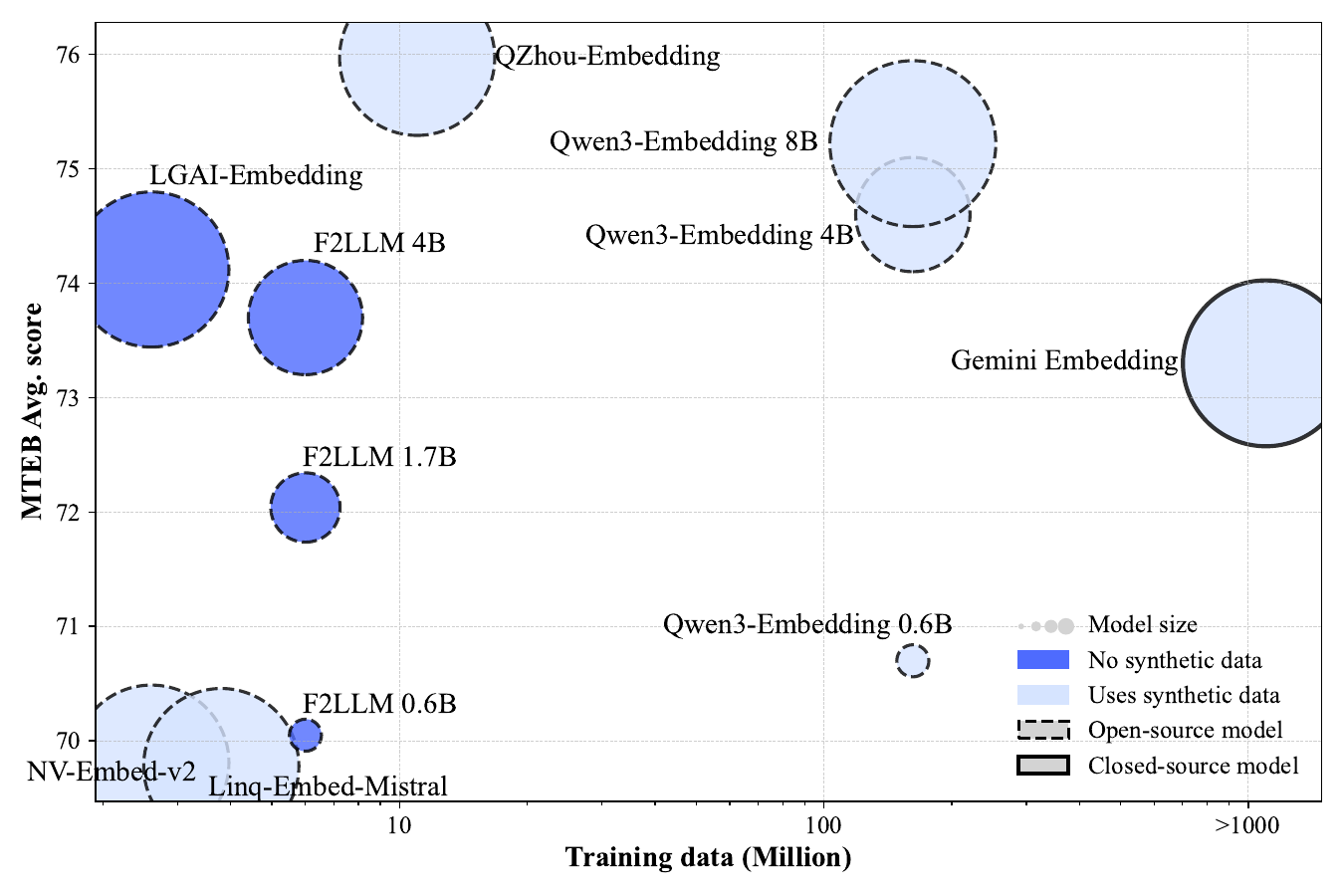}
    \includegraphics[width=0.40\linewidth]{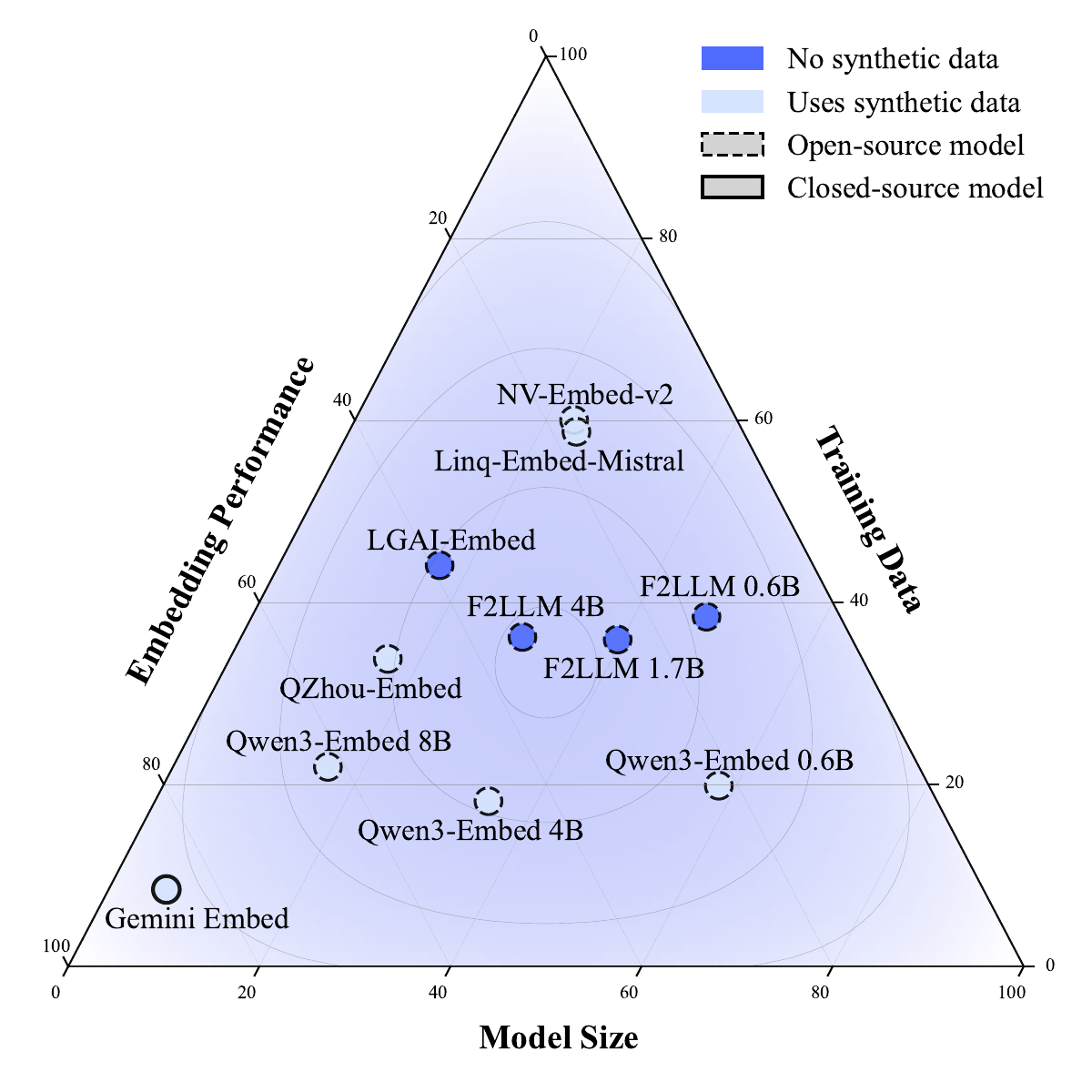}
    \caption{(Left): MTEB performance comparison between LLM-based embedding models. (Right): F2LLM, trained solely on open-source non-synthetic data, achieves a strong balance between embedding performance, training data, and model size. Higher scores indicate better performance (left axis), fewer training data (right axis), and smaller model size (bottom axis).}
    \label{fig:overview}
\end{figure}

\vspace{0.3cm}
\begin{abstract}
We introduce \textbf{F2LLM - Foundation to Feature Large Language Models}, a suite of state-of-the-art embedding models in three sizes: 0.6B, 1.7B, and 4B. Unlike previous top-ranking embedding models that require massive contrastive pretraining, sophisticated training pipelines, and costly synthetic training data, F2LLM is directly finetuned from foundation models on 6 million query-document-negative tuples curated from open-source, non-synthetic datasets, striking a strong balance between training cost, model size, and embedding performance. On the MTEB English leaderboard, F2LLM-4B ranks 2nd among models with approximately 4B parameters and 7th overall, while F2LLM-1.7B ranks 1st among models in the 1B-2B size range. To facilitate future research in the field, we release the models, training dataset, and code, positioning F2LLM as a strong, reproducible, and budget-friendly baseline for future works. 
\end{abstract}

\newpage
\section{Introduction}


In recent years, LLM-based embedding models have rapidly advanced information retrieval, clustering, classification, and other embedding-based applications, as witnessed by their increasing performance on the MTEB leaderboard~\citep{2023MTEB,2025MMTEB}. These models adapt foundation LLMs such as Mistral~\citep{2023Mistral} and Qwen3~\citep{2025Qwen3} to generate high-quality text embeddings via contrastive training on query-document pairs, efficiently utilizing their textual understanding acquired during large-scale pretraining. However, most SOTA embedding models either adopt sophisticated multi-stage training pipelines involving large-scale weakly supervised pretraining~\citep{2023GTE,2024BGE-M3,2025NV-Embed,2025Qwen3-Embedding}, or rely on costly synthetic data generated by LLMs~\citep{2024E5-Mistral,2025NV-Embed,2025Gemini-Embedding}, posing significant challenges in reproducing these models. Moreover, most of these models only open-source model checkpoints but not training script sand data, leading to inconsistencies in the literature.

To address these challenges, we introduce F2LLM, the new state-of-the-art embedding model family from Codefuse after D2LLM~\citep{2024D2LLM}. Unlike the previously mentioned models, \textbf{F2LLM is finetuned directly from a foundation models on 6 million high-quality query, document, and hard negative tuples, collected solely from open-source non-synthetic datasets} and covering a diverse range of task types. Moreover, \textbf{F2LLM is fully open-source, with model checkpoints, training data, and training code} all publicly released, providing a strong, reproducible baseline for future research in training and improving embedding models.

As of September 2025, F2LLM 4B ranks 2nd on the MTEB English leaderboard among models around 4B size, and 7th overall, only after models that are either closed-source or trained on hundreds of millions of data. Meanwhile, F2LLM 1.7B ranks 1st among models of 1B-2B size, making it an ideal choice for applications with limited computational resources.

\section{Related Work}

Adapting LLMs into embedding models has dominated the embedding literature in recent years. Most of these models adopt a two-stage training pipeline, with billion-scale weakly supervised contrastive pretraining followed by supervised finetuning~\citep{2023GTE,2024BGE-M3,2024Gecko,2024mGTE,2025Gemini-Embedding,2025Qwen3-Embedding}. A few of them, however, choose different designs, either benefiting from high-quality data or being constrained by heterogeneous data formats. For example, E5-Mistral~\citep{2024E5-Mistral} and LGAI-Embedding~\citep{2025LGAI-Embedding-Preview} are both finetuned in a single stage, while NV-Embed~\citep{2025NV-Embed} is first trained on retrieval data with both in-batch loss and hard negative loss, and then on a mixture of retrieval and non-retrieval data with only hard negative loss.

Another common characteristic of LLM-based embedding models is the usage of synthetic data. E5-Mistral~\citep{2024E5-Mistral}, Gecko~\citep{2024Gecko}, NV-Embed~\citep{2025NV-Embed}, Gemini-Embedding~\citep{2025Gemini-Embedding} and Qwen3-Embedding~\citep{2025Qwen3-Embedding} all utilize existing LLMs to generate high-quality data for supervised finetuning or even pretraining. However, recently LGAI-Embedding~\citep{2025LGAI-Embedding-Preview} achieved SOTA performance on the MTEB leaderboard, demonstrating that a carefully curated composite of open-source datasets can lead to the same performance without the expansive costs of synthetic data generation.

Finally, the increase in embedding performance often comes at the cost of increased computation or modified model architectures. Models such as BGE-ICL~\citep{2025BGE-ICL}, PromptEOL~\citep{2024PromptEOL}, and LGAI-Embedding~\citep{2025LGAI-Embedding-Preview} utilize in-context learning to improve embedding quality, while ECHO~\citep{2025ECHO} repeats the input to overcome the limitation of causal attention in LLMs. Other models, in contrast, choose to remove the causal mask~\citep{2024LLM2Vec,2025GritLM,2025NV-Embed,2025Gemini-Embedding,2025KaLM-Embedding-V2}. While this modification to attention does not increase computational cost, research has demonstrated that it does not lead to consistent performance gain either~\citep{2025BGE-ICL}.

In contrast to prior work, F2LLM is trained directly from foundation models in a single stage, without any modification to the base LLM's architecture or input format. More importantly, F2LLM is trained solely on open-source data, marking it as a reproducible and budget-friendly baseline for future works.
\section{F2LLM}

To address the problem of scattered and inconsistent training data in embedding model research, we compile a large-scale composite covering 4.9M retrieval samples, 0.2M classification samples, and 0.8M clustering samples in a unified format. For all datasets, we use the training set provided by MTEB whenever available~\citep{2023MTEB,2025MMTEB}, and otherwise perform decontamination if a corresponding test set exists in MTEB. The complete list of data sources and related statistics are given in Appendix~\ref{appendix:training-data}.

\subsection{Data Collection and Formatting}\label{sec:data-collection}

To simplify the training process of embedding models, we compile data from different types of tasks in a unified format. Each data sample consists of a (query, positive passage, hard negative $\times n$) tuple, where $n$ is set to 24 for retrieval and clustering tasks, and 1 for classification tasks. Following the practices of SOTA models, we append a task-specific instruction to each query in the following format:
\begin{equation}
    q_{\text{inst}} = \texttt{Instruct: \{task\_instruction\} }\backslash\texttt{n Query:} \{q\},
\end{equation}
where the complete list of task instructions are given in Appendix~\ref{appendix:instruction}.

\subsubsection{Retrieval Data and Margin-Based Adaptive Hard Negative Mining}

In F2LLM training data, retrieval, summarization, NLI, STS, and paraphrase (i.e. duplicate detection) datasets are collated into a unified retrieval format.

\textbf{Existing retrieval and question-answering datasets} typically come in the form of queries, passages, and a query-passage relation matrix. For each query, we sample a related passage as the positive, and mine hard negatives using the Sentence Transformers library~\citep{2013Sentence-Compression}. For each query in a dataset, we retrieve the top 100 most relevant passages using Qwen3-Embedding-0.6B~\citep{2025Qwen3-Embedding} but exclude the top 5 passages to avoid false negatives. We filter these passages to retain only those with a score below 0.8 and also less than 95\% below the score for the positive. We then select the top 24 passages in the remaining pool as hard negatives for the query, and discard the query if there are fewer than 24 passages left. \textbf{Summarization datasets} are also considered in this category, where the original text serves as the passage and its summary as the query.

For \textbf{natural language inference (NLI) datasets}, we retain only premises with at least one entailed hypothesis as queries, and sample one of the entailed hypotheses as the positive. If the query is also paired with neutral or contradictory hypotheses, we add them as hard negatives, and mine the remaining hard negatives as described above.

For \textbf{textual similarity (STS) datasets}, we follow the practice of \citet{2025NV-Embed} and construct a query-positive pair from any sentence pair whose similarity score is at least 4 and another pair with the query and positive switched. Hard negatives are also mined as described above.

Finally, for \textbf{duplicate detection datasets} (e.g. QQP~\citealp{2019QQP}), we use duplicate questions as query-positive pairs, and similarly mine hard negatives.

\subsubsection{Classification and Clustering Data}\label{sec:dataset-classification}

For \textbf{binary classification datasets}, we utilize Amazon Counterfactual~\citep{2021Amazon-Counterfactual}, Amazon Polarity~\citep{2013Amazon-Reviews}, IMDb sentiment classification~\citep{2011IMDB}, Toxic Conversations~\citep{2019Toxic-Conversations}, and CoLA~\citep{2019CoLA}. Following previous works~\citep{2025NV-Embed}, we treat each input as a query, its label text (e.g. ``toxic'') as the positive passage, and the other class's label text (e.g. ``not toxic'') as one hard negative.

For \textbf{multi-way classification and clustering datasets}, we also follow \citet{2025NV-Embed}. For each query, a random sample from the same class or cluster is used as its positive passage, and 24 samples from other classes are selected as hard negatives.


\subsection{Training}\label{sec:baseline-training}

Given a query $q_i$, its corresponding document $d_i^+$, and $n$ hard negatives $d_{i,1}^-, d_{i,2}^-,\cdots,d_{i,n}^-$, hard negative loss is computed using the contrastive learning loss
\begin{equation}
    \ell_\text{hard} = - \log\frac{e^{s(q_i,d_i^+)/\tau}}{e^{s(q_i,d_i^+)/\tau}+\sum\limits_{j=1}^{n}e^{s(q_i,d_{i,j}^-)/\tau}},
\end{equation}

where $\tau=0.05$ is the temperature, and $s()$ is a similarity score implemented as the cosine similarity between query and document embeddings.

Similarly, in-batch loss is computed by
\begin{equation}
    \ell_\text{in-batch} = - \log\frac{e^{s(q_i,d_i^+)/\tau}}{\sum\limits_{j=1}^{B}e^{s(q_i,d^+_j)/\tau}},
\end{equation}
where $B$ is the batch size.

The final loss is an unweighted sum of hard negative and in-batch loss:
\begin{equation}
    \ell = \ell_\text{hard} + \ell_\text{in-batch}.
\end{equation}

Hard negative loss is computed for all tasks. In contrast, in-batch loss is only computed for retrieval tasks. This is achieved by implementing a custom multitask dataloader, which ensures that samples in each micro batch belong to the same data source. In each training step, the dataloader on each GPU independently samples a data source with probabilities proportional to the size of the data source to ensure that all datasets finish one epoch before any dataset starts the second iteration. If a retrieval or clustering dataset is selected, 7 negatives are randomly sampled from the pool of 24 negatives for each query in the batch, while for classification datasets only one negative is used as described in Section~\ref{sec:dataset-classification}. Hard negative loss is calculated independently on each GPU for all datasets, while in-batch loss is calculated only for retrieval tasks, using all passages in the entire mini batch\footnote{We use ``micro batch'' to refer to data on a single GPU in one optimization step, and ``mini batch'' to refer to the collection of data on all GPUs in one optimization step.}
. We note that this differs from recent works that disable in-batch loss entirely after blending in non-retrieval data~\citep{2025NV-Embed}, allowing for greater sample efficiency.

\begin{table}[h]
    \centering
    \caption{Hyperparameters for training F2LLM models.}
    \begin{tabular}{rrrrrr}
    \toprule
    Model & Learning Rate & Global Batch Size & Num. GPUs & Micro Batch Size \\
    \midrule
        0.6B & 1e-5 & 512 & 16 & 32 \\
        1.7B & 9e-6 & 512 & 16 & 32 \\
        4B & 8e-6 & 512 & 32 & 16 \\
    \bottomrule
    \end{tabular}
    \label{tab:hyperparameter}
\end{table}

We use Qwen3 models~\citep{2025Qwen3} as the backbone, and directly conduct contrastive finetuning without any weakly supervised pretraining due to the large scale and high quality of our dataset. We train models of three sizes: 0.6B, 1.7B, 4B. The models are trained with AdamW optimizer~\citep{2019AdamW} and cosine learning rate decay for 2 epochs with 500 warmup steps. ZeRO stage 2~\citep{2020zero}, Flash Attention 2~\citep{2024FlashAttention2}, and gradient checkpointing are enabled to reduce GPU memory usage. Max input length is set to 1024 tokens, and the rest of the hyperparameters are given in Table~\ref{tab:hyperparameter}.

\section{Experiments}

\begin{table}[t]
    \centering
    \caption{Top models on the MTEB leaderboard. ``Average'' column is the micro average of all 41 tasks.}
    \adjustbox{width=\textwidth+0.1cm,center}{
    \begin{tabular}{m{3.9cm}>{\raggedleft\arraybackslash}m{1.2cm}>{\raggedleft\arraybackslash}m{1.3cm}>{\raggedleft\arraybackslash}m{1.3cm}>{\raggedleft\arraybackslash}m{1.3cm}>{\raggedleft\arraybackslash}m{1.3cm}>{\raggedleft\arraybackslash}m{1.3cm}>{\raggedleft\arraybackslash}m{1.3cm}>{\raggedleft\arraybackslash}m{1.3cm}r}
    \toprule
        Model & Classifi-cation & Clustering & PairClassi-fication & Reranking & Retrieval & STS & Summari-zation & \textbf{Average} & \textbf{Rank} \\
    \midrule
        \rowcolor{gray!15}
        \multicolumn{10}{c}{\emph{Closed-source}} \\
        Seed1.5-Embedding & 89.88 & 60.83 & 87.39 & 50.67 & 67.45 & 87.23 & 36.44 & 74.76 & 3 \\
        Seed1.6-Embedding & 92.42 & 59.22 & 85.07 & 50.28 & 64.90 & 86.87 & 37.10 & 74.07 & 6 \\
        Gemini Embedding & 90.05 & 59.39 & 87.70 & 48.60 & 64.35 & 85.29 & 38.28 & 73.30 & 8 \\
    \midrule
        \rowcolor{gray!15}
        \multicolumn{10}{c}{\emph{7-8B}} \\
        QZhou-Embedding 7B & 88.97 & 61.65 & 92.43 & 51.77 & 67.12 & 91.65 & 33.05 & 75.97 & 1 \\
        Qwen3-Embedding 8B & 90.43 & 58.57 & 87.52 & 51.56 & 69.44 & 88.58 & 34.83 & 75.22 & 2 \\
        LGAI-Embedding 7B & 89.97 & 59.25 & 88.67 & 49.13 & 66.18 & 86.69 & 38.93 & 74.12 & 5 \\
        GTE-Qwen2 7B & 88.52 & 58.97 & 85.90 & 50.47 & 58.09 & 82.69 & 35.74 & 70.72 & 11\\
        GeoEmbedding 7B & 89.67 & 56.50 & 82.51 & 48.32 & 60.91 & 80.60 & 30.43 & 70.21 & 13 \\
    \midrule
        \rowcolor{gray!15}
        \multicolumn{10}{c}{\emph{4B}} \\
        Qwen3-Embedding 4B & 89.84 & 57.51 & 87.01 & 50.76 & 68.46 & 88.72 & 34.39 & 74.60 & 4 \\
        F2LLM 4B & 91.68 & 68.54 & 83.75 & 50.05 & 59.63 & 84.20 & 33.19 & 73.67 & 7 \\
    \midrule
        \rowcolor{gray!15}
        \multicolumn{10}{c}{\emph{1-2B}} \\
        F2LLM 1.7B & 90.86 & 64.13 & 83.27 & 49.84 & 57.83 & 83.90 & 29.88 & 72.01 & 9 \\
        Jasper 2B & 90.27 & 60.52 & 88.14 & 50.00 & 56.05 & 84.37 & 37.19 & 71.41 & 10 \\
    \midrule
        \rowcolor{gray!15}
        \multicolumn{10}{c}{\emph{<1B}} \\
        Qwen3-Embedding 0.6B & 85.76 & 54.05 & 84.37 & 48.18 & 61.83 & 86.57 & 33.43 & 70.70 & 12 \\
        F2LLM 0.6B & 90.56 & 60.36 & 81.49 & 47.88 & 55.70 & 82.48 & 24.54 & 70.03 & 14 \\
    \bottomrule
    \end{tabular}
    }
    \label{tab:baseline-results}
\end{table}

We evaluate the models on 41 English tasks in MTEB~(\citealp{2023MTEB,2025MMTEB}, task details given in Appendix~\ref{appendix:instruction}), and compare the results with state-of-the-art embedding models on the leaderboard in Table~\ref{tab:baseline-results}.

Overall, F2LLM achieves performance comparable to SOTA embedding models trained on hundreds of millions of data samples. F2LLM-4B ranks 2nd among models of similar size and 7th overall on the leaderboard, while F2LLM-1.7B ranks 1st among models in 1B-2B size, and F2LLM-0.6B ranks 2nd among models with less than 1B parameters. Notably, F2LLM excels in clustering tasks, with the 4B model scoring 68.54, setting a new record among all models.
\section{Conclusion}

We present F2LLM, a family of fully open embedding LLMs that achieve a strong balance between model size, training data, and embedding performance. The model checkpoints, training dataset, and training code are released, positioning F2LLM as a strong, reproducible, and budget-friendly baseline for future research in text embedding models.

\bibliographystyle{colm2024_conference}
\bibliography{custom}

\appendix
\clearpage
\section{Training Data}\label{appendix:training-data}

\begin{table}[ht]
    \centering
    \caption{Classification and clustering datasets used for training F2LLM. Average query and corpus lengths are measured by the number of Qwen3 tokens, and the length of queries includes instructions.}
    \adjustbox{width=\textwidth-0.0cm,center}{
    \rowcolors{2}{gray!10}{white}
    \begin{tabular}{m{3cm}crr>{\raggedleft\arraybackslash}m{1cm}>{\raggedleft\arraybackslash}m{1cm}m{3cm}m{7cm}}
    \toprule
Dataset & Type & \# Query & Corpus Size & Query Length & Corpus Length & Source & URL \\
    \midrule
Amazon Counterfactual & Classification & 8,663 & 2 & 47 & 4 & \citet{2021Amazon-Counterfactual} & \url{https://huggingface.co/datasets/mteb/amazon_counterfactual} \\
Amazon Polarity & Classification & 100,000 & 2 & 115 & 2 & \citet{2013Amazon-Reviews} & \url{https://huggingface.co/datasets/mteb/amazon_polarity} \\
IMDb & Classification & 24,904 & 2 & 314 & 2 & \citet{2011IMDB} & \url{https://huggingface.co/datasets/mteb/imdb} \\
Toxic Conversations & Classification & 49,900 & 2 & 84 & 3 & \citet{2019Toxic-Conversations} & \url{https://huggingface.co/datasets/mteb/toxic_conversations_50k} \\
CoLA & Classification & 9,571 & 2 & 28 & 2 & \citet{2019CoLA} & \url{https://gluebenchmark.com/tasks} \\
Amazon Reviews & Clustering & 100,000 & 99,966 & 63 & 48 & \citet{2013Amazon-Reviews} & \url{https://huggingface.co/datasets/mteb/amazon_reviews_multi} \\
Banking77 & Clustering & 9,993 & 9,993 & 31 & 15 & \citet{2020Banking77} & \url{https://huggingface.co/datasets/mteb/banking77} \\
Emotion & Clustering & 17,944 & 17,944 & 54 & 21 & \citet{2018Emotion} & \url{https://huggingface.co/datasets/mteb/emotion} \\
MTOP Intent & Clustering & 17,896 & 17,862 & 28 & 9 & \citet{2021MTOP} & \url{https://huggingface.co/datasets/mteb/mtop_intent} \\
MTOP Domain & Clustering & 17,538 & 17,531 & 29 & 10 & \citet{2021MTOP} & \url{https://huggingface.co/datasets/mteb/mtop_domain} \\
Massive Scenario & Clustering & 13,547 & 13,488 & 26 & 8 & \citet{2023MASSIVE} & \url{https://huggingface.co/datasets/mteb/amazon_massive_scenario} \\
Massive Intent & Clustering & 13,547 & 13,488 & 26 & 8 & \citet{2023MASSIVE} & \url{https://huggingface.co/datasets/mteb/amazon_massive_intent} \\
Tweet Sentiment Extraction & Clustering & 26,732 & 26,732 & 40 & 19 & \citet{2020tweet-sentiment-extraction} & \url{https://huggingface.co/datasets/mteb/tweet_sentiment_extraction} \\
Arxiv-Clustering-P2P & Clustering & 83,476 & 299,733 & 245 & 222 & \citet{2022mteb-arxiv} & \url{https://huggingface.co/datasets/mteb/raw_arxiv} \\
Arxiv-Clustering-S2S & Clustering & 83,486 & 299,630 & 38 & 17 & \citet{2022mteb-arxiv} & \url{https://huggingface.co/datasets/mteb/raw_arxiv} \\
Biorxiv-Clustering-P2P & Clustering & 57,296 & 57,215 & 360 & 338 & \citet{2022mteb-biorxiv} & \url{https://huggingface.co/datasets/mteb/raw_biorxiv} \\
Biorxiv-Clustering-S2S & Clustering & 57,296 & 57,204 & 42 & 22 & \citet{2022mteb-biorxiv} & \url{https://huggingface.co/datasets/mteb/raw_biorxiv} \\
Medrxiv-Clustering-P2P & Clustering & 18,659 & 18,653 & 453 & 431 & \citet{2022mteb-medrxiv} & \url{https://huggingface.co/datasets/mteb/raw_medrxiv} \\
Medrxiv-Clustering-S2S & Clustering & 18,659 & 18,652 & 45 & 25 & \citet{2022mteb-medrxiv} & \url{https://huggingface.co/datasets/mteb/raw_medrxiv} \\
Reddit-Clustering-P2P & Clustering & 80,000 & 489,218 & 194 & 175 & \citet{2021Reddit-Clustering-P2P} & \url{https://huggingface.co/datasets/sentence-transformers/reddit-title-body} \\
Reddit-Clustering-S2S & Clustering & 58,141 & 104,725 & 34 & 15 & \citet{2021Reddit-StackExchange-Clustering-S2S} & \url{https://github.com/UKPLab/TWEAC-qa-agent-selection/tree/master/data/reddit/train} \\
StackExchange-Clustering-P2P & Clustering & 80,000 & 430,913 & 304 & 281 & \citet{2021StackExchange-Clustering-P2P} & \url{https://huggingface.co/datasets/flax-sentence-embeddings/stackexchange_title_body_jsonl} \\
StackExchange-Clustering-S2S & Clustering & 56,731 & 674,714 & 32 & 13 & \citet{2021Reddit-StackExchange-Clustering-S2S} & \url{https://github.com/UKPLab/TWEAC-qa-agent-selection/tree/master/data/stackexchange/train} \\
TwentyNewsgroups & Clustering & 11,060 & 10,994 & 230 & 216 & \citet{1995TwentyNewsGroups} & \url{https://huggingface.co/datasets/SetFit/20_newsgroups} \\
    \midrule
    \rowcolor{white} Total & Classification & 193,038 & 10 \\
    \rowcolor{white} Total & Clustering & 822,001 & 2,678,655 \\
    \bottomrule
    \end{tabular}
    }
    \label{tab:training-data-classification-clustering}
\end{table}

\begin{table}[ht]
    \centering
    \caption{Retrieval datasets used for training F2LLM. Average query and corpus lengths are measured by the number of Qwen3 tokens, and the length of queries includes instructions.}
    \adjustbox{width=\textwidth-0.0cm,center}{
    \rowcolors{2}{gray!10}{white}
    \begin{tabular}{m{3cm}crr>{\raggedleft\arraybackslash}m{1cm}>{\raggedleft\arraybackslash}m{1cm}m{3cm}m{7cm}}
    \toprule
Dataset & Type & \# Query & Corpus Size & Query Length & Corpus Length & Source & URL \\
    \midrule
Arguana & Retrieval & 22,848 & 8,561 & 25 & 210 & \citet{2018Arguana} & \url{https://huggingface.co/datasets/BeIR/arguana-generated-queries} \\
SNLI & Retrieval & 54,585 & 320,753 & 32 & 10 & \citet{2015SNLI} & \url{https://huggingface.co/datasets/stanfordnlp/snli} \\
MNLI & Retrieval & 112,075 & 358,808 & 42 & 14 & \citet{2018MNLI} & \url{https://huggingface.co/datasets/nyu-mll/multi_nli} \\
ANLI & Retrieval & 18,801 & 104,706 & 95 & 14 & \citet{2020ANLI} & \url{https://huggingface.co/datasets/facebook/anli} \\
PAQ & Retrieval & 938,771 & 2,492,657 & 30 & 145 & \citet{2021PAQ} & \url{https://huggingface.co/datasets/sentence-transformers/paq} \\
SQuAD & Retrieval & 89,509 & 20,951 & 30 & 162 & \citet{2016SQuAD} & \url{https://huggingface.co/datasets/rajpurkar/squad} \\
StackExchange & Retrieval & 754,705 & 2,401,305 & 231 & 245 & \citet{2021StackExchangeDataset} & \url{https://huggingface.co/datasets/flax-sentence-embeddings/stackexchange_titlebody_best_voted_answer_jsonl} \\
MSMARCO & Retrieval & 365,503 & 4,472,684 & 26 & 82 & \citet{2016MSMARCO} & \url{https://huggingface.co/datasets/mteb/msmarco} \\
Natural Questions & Retrieval & 97,209 & 75,178 & 30 & 145 & \citet{2019NaturalQuestions} & \url{https://huggingface.co/datasets/sentence-transformers/natural-questions} \\
HotpotQA & Retrieval & 120,528 & 1,217,525 & 45 & 103 & \citet{2018HotpotQA} & \url{https://huggingface.co/datasets/mteb/hotpotqa} \\
FEVER & Retrieval & 106,605 & 441,174 & 31 & 324 & \citet{2018FEVER} & \url{https://huggingface.co/datasets/mteb/fever} \\
ELI5 & Retrieval & 161,345 & 215,884 & 42 & 229 & \citet{2019ELI5} & \url{https://huggingface.co/datasets/Pavithree/eli5} \\
FiQA2018 & Retrieval & 7,452 & 32,615 & 33 & 234 & \citet{2018FIQA} & \url{https://huggingface.co/datasets/mteb/fiqa} \\
BioASQ & Retrieval & 125,248 & 149,900 & 27 & 295 & \citet{2015BioASQ} & \url{https://huggingface.co/datasets/BeIR/bioasq-generated-queries} \\
NFCorpus & Retrieval & 1,283 & 3,270 & 23 & 333 & \citet{2016NFCorpus} & \url{https://huggingface.co/datasets/mteb/nfcorpus} \\
MIRACL & Retrieval & 3,379 & 31,129 & 27 & 158 & \citet{2023MIRACL} & \url{https://huggingface.co/datasets/miracl/miracl} \\
Mr.TyDi & Retrieval & 3,547 & 79,695 & 27 & 148 & \citet{2021mrtidy} & \url{https://huggingface.co/datasets/mteb/mrtidy} \\
SciFact & Retrieval & 859 & 4,506 & 40 & 353 & \citet{2020SciFact} & \url{https://huggingface.co/datasets/mteb/scifact} \\
TriviaQA & Retrieval & 60,025 & 1,014,344 & 36 & 149 & \citet{2017TriviaQA} & \url{https://huggingface.co/datasets/sentence-transformers/trivia-qa-triplet} \\
COLIEE & Retrieval & 454 & 532 & 61 & 123 & \citet{2022COLIEE} & \url{https://www.modelscope.cn/datasets/sentence-transformers/coliee} \\
PubMedQA & Retrieval & 60,227 & 61,233 & 39 & 314 & \citet{2019PubMedQA} & \url{https://huggingface.co/datasets/qiaojin/PubMedQA} \\
S2ORC-Title-Abstract & Retrieval & 250,000 & 2,476,989 & 34 & 136 & \citet{2020S2ORC} & \url{https://huggingface.co/datasets/sentence-transformers/s2orc} \\
S2ORC-Title-Citation & Retrieval & 132,879 & 1,619,105 & 36 & 21 & \citet{2020S2ORC} & \url{https://huggingface.co/datasets/sentence-transformers/s2orc} \\
S2ORC-Abstract-Citation & Retrieval & 231,587 & 1,615,793 & 259 & 256 & \citet{2020S2ORC} & \url{https://huggingface.co/datasets/sentence-transformers/s2orc} \\
Amazon QA & Retrieval & 59,340 & 894,812 & 43 & 69 & \citet{2019AmazonQA} & \url{https://github.com/amazonqa/amazonqa} \\
SPECTER & Retrieval & 24,717 & 199,028 & 38 & 14 & \citet{2020SPECTER} & \url{https://huggingface.co/datasets/sentence-transformers/specter} \\
XSum & Retrieval & 184,383 & 214,562 & 44 & 459 & \citet{2018XSum} & \url{https://huggingface.co/datasets/EdinburghNLP/xsum} \\
CNN\_DM & Retrieval & 100,000 & 290,602 & 82 & 766 & \citet{2015CNN-DM} & \url{https://huggingface.co/datasets/abisee/cnn_dailymail} \\
Sentence Compression & Retrieval & 175,477 & 175,477 & 26 & 33 & \citet{2013Sentence-Compression} & \url{https://huggingface.co/datasets/sentence-transformers/sentence-compression} \\
StackExchange-DupQuestions-S2S & Retrieval & 183,559 & 158,628 & 31 & 14 & \citet{2021Embedding-Training-Data} & \url{https://huggingface.co/datasets/sentence-transformers/stackexchange-duplicates} \\
StackExchange-DupQuestions-P2P & Retrieval & 203,060 & 124,283 & 189 & 180 & \citet{2021Embedding-Training-Data} & \url{https://huggingface.co/datasets/sentence-transformers/stackexchange-duplicates} \\
QQP & Retrieval & 243,598 & 445,064 & 31 & 13 & \citet{2019QQP} & \url{https://gluebenchmark.com/tasks} \\
StackOverflow-DupQuestions & Retrieval & 19,847 & 315,577 & 27 & 11 & \citet{2018StackOverflowDupQuestions} & \url{https://huggingface.co/datasets/mteb/stackoverflowdupquestions-reranking} \\
STS12 & Retrieval & 1,858 & 2,612 & 40 & 27 & \citet{2012STS12} & \url{https://huggingface.co/datasets/mteb/sts12-sts} \\
STS22 & Retrieval & 389 & 1,380 & 505 & 494 & \citet{2022STS22} & \url{https://huggingface.co/datasets/mteb/sts22-crosslingual-sts} \\
STSBenchmark & Retrieval & 3,297 & 9,247 & 25 & 14 & \citet{2021STSBenchmark} & \url{https://huggingface.co/datasets/mteb/stsbenchmark-sts} \\
    \midrule
    \rowcolor{white} Total & Retrieval & 4,918,949 & 22,050,569 \\
    \bottomrule
    \end{tabular}
    }
    \label{tab:training-data-retrieval}
\end{table}

\clearpage
\section{Embedding Instructions}\label{appendix:instruction}

\begin{table}[ht]
    \centering
    \caption{Instructions for evaluation tasks in MTEB.}
    \adjustbox{width=\textwidth-0.8cm,center}{
    \rowcolors{2}{gray!10}{white}
    \begin{tabular}{m{4cm}m{12cm}}
    \toprule
    Task & Instruction \\
    \midrule
AmazonCounterfactual-Classification & Classify a given Amazon customer review text as either counterfactual or not counterfactual. \\
ArXivHierarchical-ClusteringP2P & Identify the main and secondary category of arXiv papers based on the titles and abstracts. \\
ArXivHierarchical-ClusteringS2S & Identify the main and secondary category of arXiv papers based on the titles. \\
ArguAna & Given a claim, find documents that refute the claim. \\
AskUbuntuDupQuestions & Retrieve duplicate questions from AskUbuntu forum. \\
BIOSSES & Retrieve semantically similar text. \\
Banking77Classification & Given an online banking query, find the corresponding intents. \\
BiorxivClusteringP2P.v2 & Identify the main category of bioRxiv papers based on the titles and abstracts. \\
CQADupstack-GamingRetrieval & Given a question, retrieve questions that are semantically equivalent. \\
CQADupstack-UnixRetrieval & Given a question, retrieve questions that are semantically equivalent. \\
ClimateFEVER-HardNegatives & Given a claim about climate change, retrieve documents that support or refute the claim. \\
FEVERHardNegatives & Given a claim, retrieve documents that support or refute the claim. \\
FiQA2018 & Given a financial question, retrieve passages that answer the question. \\
HotpotQAHardNegatives & Given a multi-hop question, retrieve passages that answer the question. \\
ImdbClassification & Classify the sentiment expressed in the given movie review text from the IMDB dataset. \\
MTOPDomainClassification & Classify the intent domain of the given utterance in task-oriented conversation. \\
MassiveIntentClassification & Given a user utterance as query, find the user intents. \\
MassiveScenario-Classification & Given a user utterance as query, find the user scenarios. \\
MedrxivClusteringP2P.v2 & Identify the main category of medRxiv papers based on the titles and abstracts. \\
MedrxivClusteringS2S.v2 & Identify the main category of medRxiv papers based on the titles. \\
MindSmallReranking & Retrieve relevant news articles based on user browsing history. \\
SCIDOCS & Given a scientific paper title, retrieve paper abstracts that are cited by the given paper. \\
SICK-R & Retrieve semantically similar text. \\
STS12, STS13, STS14, STS15, STS17, STS22.v2, STSBenchmark & Retrieve semantically similar text. \\
SprintDuplicateQuestions & Retrieve duplicate questions from Sprint forum. \\
StackExchange-Clustering.v2 & Identify the topic or theme of StackExchange posts based on the titles. \\
StackExchange-ClusteringP2P.v2 & Identify the topic or theme of StackExchange posts based on the given paragraphs. \\
SummEval-Summarization.v2 & Given a news summary, retrieve other semantically similar summaries. \\
TRECCOVID & Given a query on COVID-19, retrieve documents that answer the query. \\
Touche2020Retrieval.v3 & Given a question, retrieve passages that answer the question. \\
ToxicConversations-Classification & Classify the given comments as either toxic or not toxic. \\
TweetSentimentExtraction-Classification & Classify the sentiment of a given tweet as either positive, negative, or neutral \\
TwentyNewsgroups-Clustering.v2 & Identify the topic or theme of the given news articles. \\
TwitterSemEval2015 & Retrieve tweets that are semantically similar to the given tweet. \\
TwitterURLCorpus & Retrieve tweets that are semantically similar to the given tweet. \\
    \bottomrule
    \end{tabular}
    }
    \label{tab:mteb-instruction}
\end{table}

\begin{table}[ht]
    \centering
    \caption{Instructions for training data of F2LLM.}
    \adjustbox{width=\textwidth-0.8cm,center}{
    \rowcolors{2}{gray!10}{white}
    \begin{tabular}{m{3.5cm}lm{12cm}}
    \toprule
    Dataset & Type & Instruction \\
    \midrule
Arguana, SQuAD, BioASQ, NFCorpus, MIRACL, Mr.TyDi & Retrieval & Given a question, retrieve passages that answer the question.\\
PAQ, StackExchange, MSMARCO, Natural Questions & Retrieval & Given a web search query, retrieve relevant passages that answer the query.\\
SNLI, MNLI, ANLI & Retrieval & Given a premise, retrieve hypotheses that are entailed by the premise.\\
HotpotQA & Retrieval & Given a multi-hop question, retrieve passages that answer the question.\\
FEVER & Retrieval & Given a claim, retrieve documents that support or refute the claim.\\
ELI5 & Retrieval & Given a question from Reddit ELI5 forum, retrieve passages that answer it.\\
FiQA2018 & Retrieval & Given a financial question, retrieve passages that answer the question.\\
SciFact & Retrieval & Given a scientific claim, retrieve passages that support or refute the claim.\\
TriviaQA & Retrieval & Given a trivia question, retrieve passages that can answer it.\\
COLIEE & Retrieval & Given a legal statement, retrieve articles that support it.\\
PubMedQA & Retrieval & Given a question, retrieve paper abstracts from PubMed that can answer it.\\
S2ORC-Title-Abstract & Retrieval & Given a paper's title, retrieve the corresponding abstract.\\
S2ORC-Title-Citation & Retrieval & Given a paper's title, retrieve papers that cite it.\\
S2ORC-Abstract-Citation & Retrieval & Given a paper's abstract, retrieve abstract of papers that cite it.\\
Amazon QA & Retrieval & Given a question about a product, retrieve Amazon reviews that can help answer it.\\
SPECTER & Retrieval & Given a scientific paper title, retrieve paper titles that are cited by the given paper.\\
XSum, CNN\_DM & Retrieval & Given a news summary, retrieve the original news article.\\
Sentence Compression & Retrieval & Given a compressed sentence, retrieve the original sentence before compression.\\
QQP, StackExchange-DupQuestions-S2S, StackExchange-DupQuestions-P2P & Retrieval & Given a question, retrieve questions that are semantically equivalent.\\
StackOverflow-DupQuestions & Retrieval & Retrieve duplicate questions from StackOverflow forum.\\
STS12, STS22, STSBenchmark & Retrieval & Retrieve semantically similar text.\\
Amazon Counterfactual & Classification & Classify a given Amazon customer review text as either counterfactual or not counterfactual.\\
Amazon Polarity & Classification & Classify the given Amazon review into positive or negative sentiment.\\
IMDb & Classification & Classify the sentiment expressed in the given movie review text from the IMDB dataset.\\
Toxic Conversations & Classification & Classify the given comments as either toxic or not toxic.\\
CoLA & Classification & Classify the given sentence as linguistically acceptable or not acceptable.\\
Amazon Reviews & Clustering & Classify the given Amazon review into its appropriate rating category.\\
Banking77 & Clustering & Given an online banking query, find the corresponding intents.\\
Emotion & Clustering & Classify the emotion expressed in the given Twitter message into one of the six emotions: anger, fear, joy, love, sadness, and surprise.\\
MTOP Intent & Clustering & Classify the intent of the given utterance in task-oriented conversation.\\
MTOP Domain & Clustering & Classify the intent domain of the given utterance in task-oriented conversation.\\
Massive Scenario & Clustering & Given a user utterance as query, find the user scenarios.\\
Massive Intent & Clustering & Given a user utterance as query, find the user intents.\\
Tweet Sentiment Extraction & Clustering & Classify the sentiment of a given tweet as either positive, negative, or neutral.\\
Arxiv-Clustering-P2P & Clustering & Identify the main and secondary category of arXiv papers based on the titles and abstracts.\\
Arxiv-Clustering-S2S & Clustering & Identify the main and secondary category of arXiv papers based on the titles.\\
Biorxiv-Clustering-P2P & Clustering & Identify the main category of bioRxiv papers based on the titles and abstracts.\\
Biorxiv-Clustering-S2S & Clustering & Identify the main category of bioRxiv papers based on the titles.\\
Medrxiv-Clustering-P2P & Clustering & Identify the main category of medRxiv papers based on the titles and abstracts.\\
Medrxiv-Clustering-S2S & Clustering & Identify the main category of medRxiv papers based on the titles.\\
Reddit-Clustering-P2P & Clustering & Identify the topic or theme of Reddit posts based on the titles and posts.\\
Reddit-Clustering-S2S & Clustering & Identify the topic or theme of Reddit posts based on the titles.\\
StackExchange-Clustering-P2P & Clustering & Identify the topic or theme of StackExchange posts based on the given paragraphs.\\
StackExchange-Clustering-S2S & Clustering & Identify the topic or theme of StackExchange posts based on the titles.\\
TwentyNewsgroups & Clustering & Identify the topic or theme of the given news articles. \\
    \bottomrule
    \end{tabular}
    }
    \label{tab:training-instruction}
\end{table}

\end{document}